\documentclass[runningheads]{llncs}

\usepackage{eccv}

\usepackage{eccvabbrv}

\usepackage{graphicx}
\usepackage{booktabs}
\usepackage[linesnumbered, ruled, vlined]{algorithm2e}
\usepackage{algorithmic}
\usepackage{color}
\usepackage{rotating}
\usepackage{tabularray}
\usepackage{lipsum}
\usepackage{appendix}

\newenvironment{packeditemize}{
\begin{list}{$\bullet$}{
\setlength{\labelwidth}{8pt}
\setlength{\itemsep}{0pt}
\setlength{\leftmargin}{\labelwidth}
\addtolength{\leftmargin}{\labelsep}
\setlength{\parindent}{0pt}
\setlength{\listparindent}{\parindent}
\setlength{\parsep}{0pt}
\setlength{\topsep}{3pt}}}{\end{list}}

\newcommand{\samelineand}{\qquad}
\usepackage[accsupp]{axessibility}  % Improves PDF readability for those with disabilities.

\usepackage[pagebackref,breaklinks,colorlinks,citecolor=eccvblue]{hyperref}
\usepackage{orcidlink}

\def\Snospace~{\S{}}

\begin{document}

\newcommand{\AlgName}{\textsc{WMD}}

\title{Finding needles in a haystack: A Black-Box Approach to Invisible Watermark Detection}

\titlerunning{Black-Box Invisible Watermark Detection}

% \author{Minzhou Pan\inst{1,2}\orcidlink{0000-1111-2222-3333} \and
% Zhengting Wang\inst{2,3}\orcidlink{1111-2222-3333-4444} \and
% Xin Dong\inst{3}\orcidlink{2222--3333-4444-5555} \and
% Vikash Sehwag\inst{3}\orcidlink{2222--3333-4444-5555} \and
% Lingjuan Lyu\inst{2}\orcidlink{1111-2222-3333-4444} \and
% Xue Lin\inst{3}\orcidlink{1111-2222-3333-4444}
% }

\author{Minzhou Pan\inst{1,2} \and
Zhenting Wang\inst{2,3} \and
Xin Dong\inst{2} \\
Vikash Sehwag\inst{2} \and
Lingjuan Lyu\inst{2} \and
Xue Lin\inst{1}
}

\authorrunning{M.~Pan et al.}

% \institute{Northeastern University, Boston MA 02115, USA \and
% Springer Heidelberg, Tiergartenstr.~17, 69121 Heidelberg, Germany

\institute{$^{1}$Northeastern University \samelineand $^{2}$Sony AI \samelineand $^{3}$Rutgers University}

\maketitle

\begin{abstract}
In this paper, we propose WaterMark Detector ($\AlgName$), the first invisible watermark detection method under a black-box and annotation-free setting. $\AlgName$ is capable of detecting arbitrary watermarks within a given detection dataset using a clean non-watermarked dataset as a reference, without relying on specific decoding methods or prior knowledge of the watermarking techniques. We develop WMD using foundations of offset learning, where a clean non-watermarked dataset enables us to isolate the influence of only watermarked samples in the reference dataset. Our comprehensive evaluations demonstrate the effectiveness of $\AlgName$, significantly outperforming naive detection methods, which only yield AUC scores around 0.5. In contrast, $\AlgName$ consistently achieves impressive detection AUC scores, surpassing 0.9 in most single-watermark datasets and exceeding 0.7 in more challenging multi-watermark scenarios across diverse datasets and watermarking methods. As invisible watermarks become increasingly prevalent, while specific decoding techniques remain undisclosed, our approach provides a versatile solution and establishes a path toward increasing accountability,  transparency, and trust in our digital visual content.

  % \keywords{Watermark Detection \and Black-box Detection \and IP Protection}
\end{abstract}

\section{Introduction}
\label{sec:intro}

For a long time, invisible digital image watermarks have served as a reliable solution for tracing plagiarism and unauthorized copying, safeguarding intellectual property rights without compromising image quality~\cite{zhang2021brief, byrnes2021data, Singh_Singh_2022, zhong2023deep}. Moreover, with the advent of generative models, such watermarks have been proposed as a means of identifying and sourcing AI-generated images~\cite{fernandez2023stable, wen2023tree}. However, the blind detection of invisible watermarks in a given image dataset, without access to the corresponding decoding algorithms, poses significant challenges.

The inherent invisibility of watermarks makes manual screening of datasets an impractical task. Moreover, the wide variety of watermarking methods~\cite{4554423, lsb2010lsbwatermark, zhu2018hidden, tancik2020stegastamp, wen2023tree, fernandez2023stable}, each employing different embedding techniques, complicates the development of a generalized Deep Neural Network (DNN) detector. Some watermarking methods are even black-box~\cite{dalle3wm}, lacking APIs for third-party users, further hindering the inclusion of all methods in the training process. Related techniques, such as Out-of-Distribution (OOD) detection and anomaly detection, also struggle to effectively identify watermarks due to the subtle perturbations they introduce, as discussed in \autoref{sec:wm_detection}.
% \vspace{0.5em}
\begin{figure*}[!t]
    \centering
    %\vspace{-2.1em}
    \includegraphics[width=\textwidth]{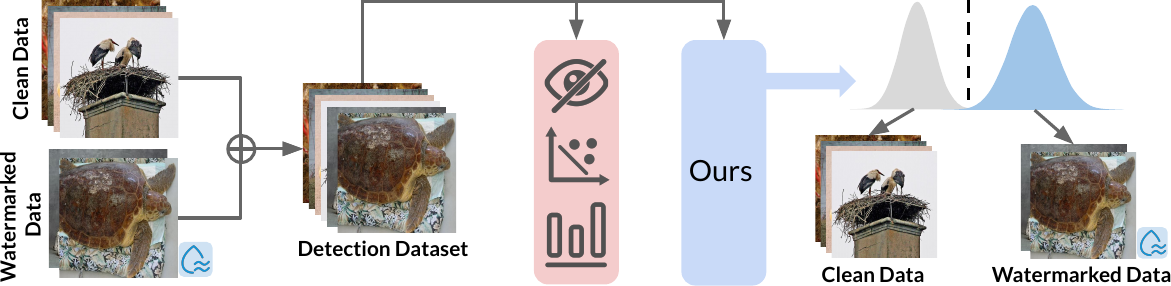}
    \vspace{-1.5em}
    \caption{\textbf{Detecting invisible watermarked in a given dataset.} Due to the invisibility of watermarks, human inspection and existing anomaly detection methods fail to distinguish watermarked images from clean ones within a dataset. To address this challenge, we propose $\AlgName$ as the first invisible watermark detection capable of accurately identifying invisible watermarked samples in the black-box setting, where there is no need for prior knowledge of the watermarking techniques or decoding methods.}
    \vspace{-1.9em}
    \label{fig:thumbnail}
\end{figure*}

Failing to detect watermarks in a dataset can lead to severe consequences. Watermarked images may contain sensitive or copyrighted information, and generative models trained on such data may inadvertently memorize these images illegally~\cite{carlini2023extracting}. Furthermore, recent studies have shown that the inclusion of watermarked images, particularly those generated by AI, in training datasets can degrade the performance of downstream models~\cite{alemohammad2023selfconsuming}. As policies increasingly mandate the use of watermarks in generative models~\cite{watermakr_law_1, watermakr_law_2, watermakr_law_3}, and with the rapid proliferation of these models~\cite{Photutorial_2023, aistatist_2023}, watermarked images are expected to become more prevalent in the near future.

In response to these emerging threats and the growing importance of watermarks in the AI-generated content landscape, we introduce the first black-box invisible watermark detection method: \textbf{W}ater\textbf{M}ark \textbf{D}etection ($\AlgName$), a method for reliably detecting arbitrary watermarks in datasets, as depicted in \autoref{fig:thumbnail}. Instead of relying on specific decoding methods for each watermark, $\AlgName$ stands out as a versatile black-box approach that eliminates the need for prior knowledge of watermarking or decoding methods. By leveraging the similar distribution of clean datasets, $\AlgName$ employs self-supervised learning to effectively identify watermarks. Extensive evaluation shows the effectiveness of $\AlgName$, consistently achieving detection AUC above 0.9 in most single-watermark datasets and above 0.7 in more challenging multi-watermark scenarios.

In this paper, we introduce $\AlgName$, the first black-box invisible watermark detection method capable of reliably identifying arbitrary watermarks in datasets without prior knowledge of watermarking or decoding techniques, leveraging self-supervised learning to exploit the similar distribution of clean datasets.

\section{Background and Related Work}

\subsection{Importance of Detecting Watermarks}
\label{sec:watermark_ratio}
Invisible digital watermarks were initially designed to protect intellectual property and copyrights without compromising the visual quality of images. Such watermarks have been widely implemented across various domains, becoming a popular solution for content sourcing~\cite{Singh_Singh_2022, byrnes2021data, Singh_Singh_2022, zhong2023deep}. Recently, watermarks have become increasingly important with the development of generative models, particularly diffusion models, which can produce photo-realistic images that are challenging for humans to distinguish from real photos~\cite{ho2020denoising, song2020denoising}. The ability to detect these watermarks is crucial for several reasons:

\noindent \textbf{Protecting Intellectual Property:} Watermarks have long been used to trace intellectual property infringement. Detecting these watermarks is essential for ensuring that copyrighted content is not collected in model training datasets and for keeping downstream models from learning this information (models can remember these details, as revealed by certain attacks~\cite{carlini2023extracting}).

\noindent \textbf{Preventing Misuse:} Generative models can be used to create fake news, propaganda, and other malicious content~\cite{lu2023seeing, Cheetham_2023}. As multiple legislation and executive orders~\cite{watermakr_law_1, watermakr_law_2, watermakr_law_3} have been proposed to require watermarks on AI-generated images, detecting watermarks can help identify AI-generated images and prevent their misuse.

\noindent \textbf{Maintaining Dataset Quality:} According to recent surveys~\cite{Photutorial_2023, aistatist_2023}, AI-generated images account for a significant and growing proportion of all images produced. Using these images in training datasets can introduce biases and inaccuracies that negatively impact the performance of downstream models~\cite{alemohammad2023selfconsuming}. Detecting and filtering out watermarked images is crucial for maintaining dataset quality and ensuring accurate model development.

The increasing prevalence of AI-generated content has led to a substantial rise in the number of watermarked images in circulation. Recent surveys~\cite{Photutorial_2023, aistatist_2023} indicate that AI-generated images account for a considerable portion of all images produced, with over 18 billion AI-generated images created within a year, and this number is growing rapidly. In comparison, human-generated images in the same period amount to around 355 billion~\cite{villalobos2022will}. Based on these statistics, we can estimate that approximately 5\% of all images created from now on will be AI-generated and potentially watermarked. This proportion will be used as a basis for subsequent evaluation in \autoref{sec:evaluation}. This trend underscores the critical importance of developing robust watermark detection methods to protect intellectual property, prevent misuse, ensure regulatory compliance, and maintain dataset quality.

\subsection{Invisible Image Watermarks}
\label{sec:watermark_background}
Invisible image watermarking embeds non-visible markers into digital images to protect copyrights and identify sources. The primary objective of such watermarking is to ensure that these markers can be readily detected by a pre-designed method while remaining imperceptible to other detection attempts and during normal use. Furthermore, the watermark should be robust and resilient to image modifications and regenerations, enabling its creator to detect it even after the image has been altered or recreated.

Various watermarking methods have been proposed to embed watermarks into images. These methods can be divided into two categories: Post-processing Watermarks and Generative Watermarks.

\noindent \textbf{Post-processing Watermarks.} This watermarking category involves the integration of watermarks directly into the image content. Traditional methods include embedding secret information into the least significant bit (LSB)~\cite{lsb2010lsbwatermark}, or incorporating watermarks within the frequency domain via transforms such as Discrete Cosine Transform (DCT) and Discrete Wavelet Transform (DWT)~\cite{465537, o1997rotation}. Other techniques exploit image decompositions, such as Singular Value Decomposition (SVD)~\cite{chang2005svd}, or employ composite transforms~\cite{4554423}.

With the advent of deep learning, new approaches to learn watermarking have emerged. Many of these methods adopt an encoder-decoder architecture, where the encoder embeds the watermark, and the decoder is responsible for its extraction~\cite{zhu2018hidden, tancik2020stegastamp}. The training of the encoder is enhanced by introducing simulated differentiable image distortions, which are placed between the encoder and decoder. This process endows the encoder with resilience against real-world image perturbations, significantly improving the robustness of the watermark. Moreover, the encoder's training regimen includes loss functions designed to minimize watermark visibility, thus advancing the limits of robustness and imperceptibility beyond what is achievable with conventional hand-designed transforms.

\noindent \textbf{Generative Watermarks.} 
Besides embedding watermarks into existing images, watermarking can be incorporated directly into the image generation process of generative models. A notable example is Stable Signature~\cite{fernandez2023stable}, which involves the training of an encoder-decoder framework. In this method, the latent decoder of the stable diffusion model is fine-tuned to act as a watermark encoder, embedding an imperceptible watermark into the generated images for subsequent detection and identification.

Another innovative approach, Tree-Ring watermarks~\cite{wen2023tree}, adapts traditional frequency-domain watermarking techniques. It leverages the Fast Fourier Transform (FFT) to transpose the diffusion latent space into a space amenable to watermarking. Therein, a unique watermark is embedded. For detection, the watermarked images undergo an inverse process through the same diffusion network and FFT, allowing for the watermark region to be cross-referenced with the original watermark for verification.

\subsection{Watermark Detection}
\label{sec:wm_detection}
As discussed in \autoref{sec:watermark_ratio}, invisible watermarks play a crucial role in protecting intellectual property rights, preventing the misuse of AI-generated content, and maintaining the quality of image datasets. However, despite the long history and diversity of invisible watermarking methods, detection techniques have not yet emerged. This limitation can be attributed to several factors:

\noindent \textbf{Infeasibility of Human Annotation.} Due to the invisibility of watermarks, it is practically impossible for humans to identify and annotate watermarked images within large-scale datasets. Consequently, the approach of manually labeling a small subset of watermarked data and training a model to detect the remaining watermarks, as employed in visible watermark detection~\cite{santoyo2017automatic, cheng2018large}, is not applicable to invisible watermarks.

\noindent \textbf{Challenges in Collecting Comprehensive Watermarking Methods.} As mentioned in \autoref{sec:watermark_background}, there exists a wide variety of watermarking methods, each with its specific embedding and decoding techniques. Even if it is possible to collect and generate watermarked datasets using a subset of these methods, it would be extremely difficult to encompass all known watermarking techniques. Moreover, many watermarking methods are proprietary and black-box~\cite{dalle3wm}, making it infeasible for ordinary users to obtain access and generate corresponding datasets.

\noindent \textbf{Limitations of Self-Supervised Approaches.} Since the label information is unavailable, several self-supervised approaches like Out-of-Distribution (OOD) detection, anomaly detection, and backdoor detection methods~\cite{pan2023asset, qi2023towards} have been proposed as substitutes to identify "abnormal" examples within a given dataset. However, these approaches fail to detect watermarked examples effectively because the perturbations introduced by watermarks are relatively small compared to typical anomalous features. Furthermore, watermarks do not cause obvious changes in model behavior, unlike backdoor examples.

To validate these findings, we conduct an experiment using the well-known DctDwtSvd~\cite{4554423} invisible watermarking method to embed watermarks in 5\% of the samples (as stated in \autoref{sec:watermark_ratio}) within a 10,000-sample subset of ImageNet~\cite{imagenet2009}. The results, summarized in \autoref{tab:othermethodfail}, demonstrate the ineffectiveness of various existing detection methods in identifying even the most basic invisible watermarks. This highlights the need for novel and robust watermark detection techniques that can overcome the limitations of current approaches.

\begin{table}[]
\centering
\vspace{-1.5em}
\caption{Performance comparison of various detection methods on an ImageNet subset containing DctDwtSvd watermarks. The low AUC scores indicate the inability of these methods to effectively detect even the most basic invisible watermarks.}
\vspace{-1.15em}
\label{tab:othermethodfail}
\resizebox{0.95\columnwidth}{!}{%
\begin{tabular}{ccccccccc}
\toprule
 & \multicolumn{2}{c}{\textbf{Visible Watermark Detection}} & & \multicolumn{2}{c}{\textbf{Anomaly/OOD Detection}} & & \multicolumn{2}{c}{\textbf{Backdoor Samples Detection}} \\ \cmidrule{2-3} \cmidrule{5-6} \cmidrule{8-9}
 & TV-L1~\cite{santoyo2017automatic} & LSW~\cite{cheng2018large} & & DROC~\cite{sohn2020learning} & RIAD~\cite{zavrtanik2021reconstruction} & & CT~\cite{qi2023towards} & ASSET~\cite{pan2023asset} \\
\midrule
\textbf{AUC ($\uparrow$)} & 0.508 & 0.512  & & 0.513 & 0.522 & & 0.514 & 0.518 \\
\bottomrule
\end{tabular}%
}
\vspace{-1.5em}
\end{table}

Considering the growing importance of invisible watermarks and the challenges associated with their detection, an effective and reliable invisible watermark detection method is crucial. To address this need, we propose $\AlgName$ (Watermark Detection), a novel self-supervised learning approach capable of successfully detecting watermarks in the given dataset with high probability without prior knowledge of the specific watermarking algorithm.

\section{Method}

\subsection{Problem Setup}
\label{sec:threat_model}
In this section, we present the threat model for our proposed invisible watermark detection method. Let \(\boldsymbol{x}^{d}_i \in \mathcal{D}_{d}\) be the dataset of images awaiting watermark detection, where some images \(\boldsymbol{x}^{d}_i\) may be watermarked by any existing invisible watermarking technique. The watermarked portion of the dataset is denoted as \(\boldsymbol{x}^{w}_i \in \mathcal{D}_{d}^w\), and the clean images are denoted as \(\boldsymbol{x}^{c}_i \in \mathcal{D}_{d}^c\). The detection dataset can also be expressed as \(\mathcal{D}_{d} = \mathcal{D}_{d}^w \cup \mathcal{D}_{d}^c\). 

\noindent \textit{Objective.} The objective of our watermark detector is to find a watermark detection method $f(.)$ that reliably identifies the watermarked images within the given dataset, such that $f(\mathcal{D}_d) \rightarrow \mathcal{D}_{d}^w$. In particular, the watermark detector aims to classify each image in $\mathcal{D}_{d}$ as watermarked or non-watermarked.

\noindent We assume that the watermark detector has no prior knowledge of the watermarking process including which images are watermarked and the type of watermark used, thus referring to the process as \textit{black-box watermark detection}. However, the detector has access to the visual distribution of images in the detection dataset \(\mathcal{D}_{w}\) and can obtain a clean dataset \(\boldsymbol{x}^{c}_i \in \mathcal{D}_{c}\) that has similar visual distribution.

The watermark detector has full access to both the clean dataset $\mathcal{D}_c$ and the detection dataset $\mathcal{D}_d$ and is allowed to use these datasets to develop the detector. We further discuss the impact of our design choices in the development of the detection in \autoref{sec:oracle_detection} and \autoref{sec:methology}.

\subsection{Oracle Watermark Detection}
\label{sec:oracle_detection}
We formulate the watermark detection problem as an offset optimization problem~\cite{pan2023asset}. Offset optimization is a technique that identifies differences between two datasets by effectively canceling out the common elements.
Consider the oracle detection case where the clean images in the detection dataset $\mathcal{D}_d$ and the clean dataset $\mathcal{D}_c$ have identical distributions, and the size of the two datasets is same and the number of watermarks is non-zero, $ N = \left | \mathcal{D}_{d} \right |= \left |\mathcal{D}_{c} \right | > \left |\mathcal{D}_{d}^c \right| >> \left |\mathcal{D}_{d}^w \right |$. With this knowledge, we can initialize a deep neural network (DNN) model, $f(\cdot;\theta)$, and calculate the gradients of the loss function $\mathcal{L}$ with respect to the model parameters $\theta$ for each dataset:

\begin{small}
\begin{equation}
\Delta\theta_{c} = \nabla_{\theta} \frac{1}{N} \sum_{\boldsymbol{x}^{c}_i \in \mathcal{D}_{c}} \mathcal{L}(f({x}^{c}_i; \theta))
\end{equation}
\vspace{-0.5em}
\begin{equation}
\Delta\theta_{d} = \nabla_{\theta} \left(\frac{1}{N} \sum_{\boldsymbol{x}^{c}_i \in \mathcal{D}_{d}^{c}} \mathcal{L}(f({x}^{c}_i; \theta)) + \frac{1}{N} \sum_{\boldsymbol{x}^{w}_i \in \mathcal{D}_{d}^{w}} \mathcal{L}(f({x}^{w}_i; \theta)) \right)
\end{equation}
\end{small}

\noindent We then get the total gradient by subtracting the gradients from the two datasets: $\Delta\theta = \Delta\theta_c - \Delta\theta_d$.

\begin{small}
\vspace{-1em}
\begin{equation}
\Delta\theta =\nabla_{\theta} \Bigg(  \frac{1}{N} \sum_{\boldsymbol{x}^{c}_i \in \mathcal{D}_{c}} \mathcal{L}(f({x}^{c}_i; \theta)) -  \frac{1}{N} \sum_{\boldsymbol{x}^{c}_i \in \mathcal{D}_{d}^{c}} \mathcal{L}(f({x}^{c}_i; \theta)) - \frac{1}{N} \sum_{\boldsymbol{x}^{w}_i \in \mathcal{D}_{d}^{w}} \mathcal{L}(f({x}^{w}_i; \theta)) \Bigg)
\end{equation}
\end{small}

\noindent Under the Oracle assumption, the clean samples from both datasets have identical distributions and there are more numbers in the clean data set, so the gradient of $\mathcal{D}_{d}^{c}$ will be cancelled:

\begin{small}
\begin{equation}
\Delta\theta = \nabla_{\theta}  \frac{1}{N} \sum_{\boldsymbol{x}^{c}_i \in (\mathcal{D}_{c} - \mathcal{D}_{d}^{c})} \mathcal{L}(f({x}^{c}_i; \theta))
-  \nabla_{\theta}  \frac{1}{N} \sum_{\boldsymbol{x}^{w}_i \in \mathcal{D}_{d}^{w}} \mathcal{L}(f({x}^{w}_i; \theta))
\end{equation}
\end{small}

\noindent Optimizing the model parameter by descent this gradient, the problem then becomes:

\vspace{-1em}
\begin{equation}
\theta^* = \arg \min_{\theta} \frac{1}{N} \sum_{\boldsymbol{x}^{c}_i \in (\mathcal{D}_{c} - \mathcal{D}_{d}^{c})} \mathcal{L}(f(\boldsymbol{x}^{c}_i; \theta)) + \arg \max_{\theta} \frac{1}{N} \sum_{\boldsymbol{x}^{w}_i \in \mathcal{D}_{d}^{w}} \mathcal{L}(f(\boldsymbol{x}^{w}_i; \theta)).
\end{equation}

\noindent The optimized model $f(\cdot;\theta^*)$ will generate high output values for watermarked samples and low values for clean samples, enabling watermark detection based on the difference between model outputs.

This analysis reveals that the performance of the oracle detector heavily depends on the \textit{similarity between the clean distributions of the two datasets and the number of watermarked samples in the detection dataset}. In practice, finding a clean dataset that perfectly offsets the gradient is challenging. To address this issue, we propose our method $\AlgName$, which relaxes the problem and enables practical watermark detection in real-world scenarios.

\subsection{$\AlgName$: Our Proposed Black-box Watermark Detector}
\label{sec:methology}

% \zw{I suggest to add a brief overall introduction to our method here.}
To address watermark detection in real-world scenarios, where finding a perfect clean dataset corresponding to the detection dataset is challenging. To this end, we propose $\AlgName$. The key components of $\AlgName$ are an asymmetric loss function and the iterative pruning strategy for the detection dataset. The following subsections provide a detailed design and analysis of each component of $\AlgName$:

\subsubsection{Asymmetric Loss.}
Considering the high similarity between watermark detection and binary classification tasks, a straightforward design approach would be to use a symmetric loss. Symmetric loss employs the same loss function for both minimization and maximization objectives, that is minimizing the model output of samples in the clean dataset and maximization the model output of samples in the detection dataset. However, our ablation study in Appendix~\autoref{sec:ablation} reveals that symmetric loss functions fail to generate satisfactory results. The underlying reason for this is that symmetric loss functions have the same loss scale for both minimization and maximization goals, leading to an automatic balance between the two objectives. If the maximize loss is smaller, the optimization will shift its focus to minimization, and vice versa. 

For the clean samples in the detection dataset, $\mathcal{D}_{d}^{c}$, its gradients are always offset by the clean samples in the clean dataset. Consequently, it is more challenging to maximize the output for $\mathcal{D}_{d}^{c}$ compared to the watermarked samples, $\mathcal{D}_{d}^{w}$, resulting in $\mathcal{D}_{d}^{c}$ consistently generating higher loss than $\mathcal{D}_{d}^{w}$. Furthermore, watermarked samples only make up a small portion of the detection dataset, meaning their total loss is already very small compared to the clean samples. The combination of these two factors causes the model to focus on optimizing the clean samples, $\mathcal{D}_{d}^{c}$, while neglecting the maximization of watermarked samples.

The evaluation results in Appendix~\autoref{sec:ablation}, \autoref{tab:loss_design} provides further support for this explanation. The diagonal of the table presents the results of using symmetric loss functions, including symmetric exponential loss, symmetric softmax loss, and symmetric BCE loss. All of these symmetric loss functions yield worse results compared to a symmetric linear loss. The reason for this is that these three loss functions will amplify the differences between losses, causing the model to focus more on the hard samples by emphasizing the high-loss samples (clean images) and downplaying the low-loss samples (watermarked images), thereby diverting attention away from the actual target. This observation confirms our previous conjecture about the limitations of symmetric loss functions in this context.

Following the analysis, we've identified the distinct characteristics of watermark detection compared to standard binary classification. As a result, we propose the asymmetric loss function to improve the detection performance.

For the clean dataset, considering the fact that all images are from trusted sources and therefore clean, we can encourage the model to pay more attention to hard examples, i.e., the samples that yield higher loss values. A common solution is to use an exponential loss function:
\vspace{-0.3em}
\begin{small}
\begin{equation}
\mathcal{L}_{exp}=\exp(f({x}^{c}_i; \theta)/\tau)
\end{equation}
\end{small}
\noindent Here, $\tau$ is a temperature scaler that assigns higher loss values to samples with higher model outputs. This focuses the model on minimizing the loss for these hard examples, ensuring that all samples in the clean dataset are strictly minimized.

For the detection dataset, which contains both clean and watermarked samples, we want to ensure that the watermarked samples are always maximized. In other words, the model should give nearly equal focus to all examples, regardless of their output. To achieve this, we can use a linear loss:
\begin{small}
\begin{equation}
\mathcal{L}_{lin}=-f({x}^{d}_i; \theta)
\end{equation}
\end{small}
\noindent Using a linear loss has two benefits. First, it ensures that the model gives almost the same weight to all examples, keeping the focus on maximizing the watermarked examples. Second, it generates lower gradients compared to the exponential loss used for clean examples, ensuring that the losses of the clean examples are strictly bound by minimization.

However, simply combining the exponential loss with the linear loss results in different scales, making it difficult to find a balancing factor between the two. To address this, we modify the exponential loss into a softmax loss:
\vspace{-0.3em}
\begin{small}
\begin{equation}
\mathcal{L}_{sm}=\log(\exp(f({x}^{c}_i; \theta)/\tau))\cdot\tau
\end{equation}
\end{small}
The overall loss then becomes:
\vspace{-0.3em}
\begin{small}
\begin{equation}
\mathcal{L}_{total}=\log(\exp(f({x}^{c}_i; \theta)/\tau))\cdot\tau - f({x}^{d}_i; \theta)
\end{equation}
\end{small}

By using this asymmetric loss design, we can effectively optimize the model to detect watermarks while minimizing the impact of clean samples on the detection performance.

\subsubsection{Iteration Pruning.}

\begin{figure*}[!ht]
    \centering
    \vspace{-0.5em}
    \includegraphics[width=0.9\textwidth]{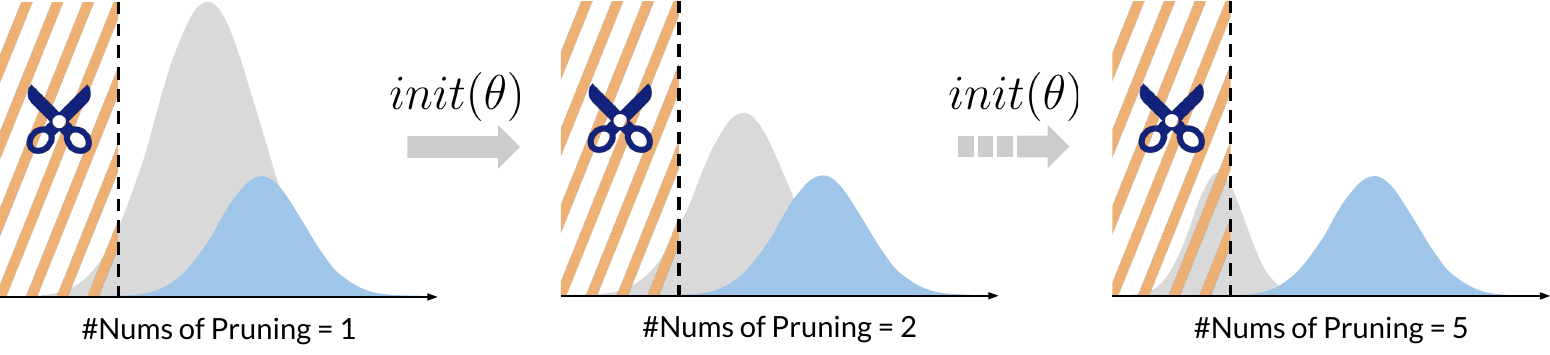}
    \vspace{-1em}
    \caption{\textbf{Illustration of the Iterative Pruning process.} As the number of pruning iterations increases, the detection dataset is gradually condensed by removing \scalebox{0.9}{\colorbox[HTML]{d9d9d9}{\textbf{clean samples}}} while retaining most of the \scalebox{0.9}{\colorbox[HTML]{9fc5e8}{\textbf{watermarked samples}}}.}
    \vspace{-2.0em}
    \label{fig:iter_pruning}
\end{figure*}

In the detection dataset, the number of clean images often significantly exceeds the number of watermarked images, making the separation of model outputs challenging. As the detection model $f(\cdot;\theta)$ is an over-parameterized deep neural network (DNN)~\cite{arpit2017closer}, it may eventually "memorize" all samples, leading to high loss values for both clean and watermarked images in the detection dataset and causing detection failure. Conversely, insufficient training results in poor separation between the clean and watermarked parts.

However, we observe that even when the model cannot perfectly separate these samples, the output values of the watermarked samples are consistently higher than those of some clean samples. This may be attributed to the fact that the watermark, even if slight, still has a different distribution compared to clean images. Leveraging this observation, we propose a strategy called "Iteration Pruning" to improve the detectability of watermarks.

The Iteration Pruning strategy is designed to efficiently refine the detection dataset and accelerate the learning process of the watermark detection model. This strategy involves two key hyperparameters: the pruning rate $\rho$ and the pruning interval $interval$. The training process begins with the initial dataset, and at every $interval$ training epochs, a percentage of data equal to $\rho$ is removed from the detection dataset. This pruning targets the samples with the lowest loss values, effectively discarding the least informative or most easily learned samples. To prevent overfitting, the model is reinitialized after each pruning step. The pruning process continues iteratively throughout the training progress until only 5\% of the total data remains. This iterative pruning approach effectively "condenses" the dataset by retaining the majority of the watermarked samples while progressively eliminating more clean samples. As the "condensing" process progresses, the model gradually focuses its attention on the watermarked samples, enabling it to quickly learn the optimal parameters for perfect separation between watermarked and clean data. 

The effectiveness of Iteration Pruning and the impact of pruning hyperparameters are investigated through an ablation study in Appendix~\autoref{sec:ablation}, providing insights into the optimal configuration for enhancing watermark detection performance.

\subsection{Overall Method}

\begin{algorithm}[tb]
\caption{Algorithm for training $\AlgName$}
\label{algo:AlgoCompact}
\DontPrintSemicolon  % Assuming you're using algorithm2e and want to avoid printing semicolons
\SetNoFillComment
\SetKwInput{KwIn}{Input}
\SetKwInput{KwOut}{Output}
\SetKwInput{KwParam}{Parameters}
% \algsetup{linenosize=\tiny} % This command depends on the package you're using. Adjust or remove accordingly.
\small

\KwIn{$\theta$ (Model); $E$ (Epochs); $\mathcal{D}_d$ (Detection Data); $\mathcal{D}_c$ (Clean Data)}
\KwOut{$\mathcal{D}_w$ (Watermarked Dataset)}
\KwParam{$\eta$ (Learning Rate), $\rho$ (Pruning Rate), $interval$ (Pruning Interval)}

$\mathcal{D}_{dt} \gets \mathcal{D}_d$

\For{$e = 1$ \KwTo $E$}{
    \For{each $N$-sized batch $\mathcal{B}_c, \mathcal{B}_{dt}$ from $\mathcal{D}_c, \mathcal{D}_{dt}$}{
        $p_c, p_{dt} \gets f(\mathcal{B}_c; \theta), f(\mathcal{B}_{dt}; \theta)$ \tcp*{Get model outputs}
        $\mathcal{L}_{total} \gets \mathcal{L}_{sm}(p_c) + \mathcal{L}_{lin}(p_{dt})$ \tcp*{Calculate the losses}
        $\theta \gets \theta - \eta \nabla_{\theta}\mathcal{L}_{total}$ \tcp*{Update the model with gradients}
    }
    \If{$e \mod interval = 0$}{
        % Prune and reinitialize steps
        
        $(p_d)\text{Ranked} \gets rank(f(\mathcal{D}_d; \theta))$ \tcp*{Get model outputs and rank}
        \tcp{Update $\mathcal{D}_{dt}$ by pruning samples with lowest outputs}
        $\mathcal{D}_{dt} \gets (p_d)\text{Ranked}[:\lfloor |\mathcal{D}_d| \times (1-(1-\rho)^{\frac{e}{interval}}) \rfloor]$ \\
        $\theta \gets init(\theta)$ \tcp*{Reinitialize model to avoid overfitting}
    }
}
\Return $\mathcal{D}_{dt}$
\end{algorithm}

The overall workflow of $\AlgName$ is as follows: First, the model is initialized with the detection dataset $\mathcal{D}_d$ and a clean dataset $\mathcal{D}_c$. During each training epoch, mini-batches from both datasets are fed into the model, and the asymmetric loss function is calculated. The model parameters are then updated using the gradients computed from the total loss. After a fixed number of epochs (determined by the $interval$ parameter), the iterative pruning process is triggered. The model's performance on the detection dataset is evaluated, and the samples are ranked based on their confidence scores. The bottom $(1-(1-\rho)^{\frac{e}{interval}})$ fraction of the ranked samples, which are most confidently predicted as non-watermarked, are removed from the detection dataset. The model is then reinitialized, and the training process continues with the updated detection dataset. This iterative pruning strategy is repeated until the specified number of epochs is reached. Finally, the detection dataset $\mathcal{D}_w$, containing the identified watermarked images, is returned.

\section{Evaluation}
\label{sec:evaluation}
In this section, we provide a comprehensive evaluation of our proposed method, $\AlgName$, for detecting invisible watermarks in a given dataset.

\subsection{Setting}

\noindent \textbf{Evaluation metrics.} As our method focuses on detecting the presence of watermarks and outputs a binary result, either ``watermarked'' or ``non-watermarked'', we employ three widely used binary detection metrics: \textit{Area Under the Curve} ({AUC}): Assesses the watermark system's discernment between watermarked and non-watermarked images, with higher AUC reflecting greater detected performance.
\textit{True Positive Rate at 10\% False Positive Rate} ({TPR @ 10\% FPR}): Demonstrates watermark detection capabilities at the expense of a small number of clean samples.
\textit{False Positive Rate at 10\% True Positive Rate} ({FPR @ 90\% TPR}): Demonstrates the false positive rate for clean images when detecting most watermarked images.

\noindent \textbf{Baseline Watermark.} We selected representative works among different watermarking methods as baseline watermarking methods to evaluate our watermark detection performance. For post-processing watermarking, we choose the least significant bit (LSB)~\cite{lsb2010lsbwatermark}, which embeds the watermark into the image's lowest bit, and DctDwtSVD (DDS)~\cite{4554423}, integrating the watermark into the DCT space and SVD vectors. Within the deep learning paradigm, HiDDeN~\cite{zhu2018hidden} utilizes an encoder-decoder architecture with a noise simulation layer for embedding, whereas StegaStamp (SS)~\cite{tancik2020stegastamp} enhances robustness by incorporating higher noise levels and improving model structure. For generative watermarking, Stable-Signature (SSig)~\cite{fernandez2023stable} embeds watermarks into latent diffusion models by fine-tuning the last layer, and Tree-Ring watermark (TR)~\cite{wen2023tree} embeds the watermark into the diffusion model latent frequency space. All the watermarks embedded 64 bits of information into the image, for the Tree-Ring watermark, we use the $\text{Tree-Ring}_{\textbf{Rings}}$ variant.

\noindent \textbf{Datasets \& Models.} 
In the main evaluation, we use three datasets as our evaluation datasets for post-processing watermarks: ImageNet~\cite{imagenet2009}, COCO~\cite{lin2014coco}, and Caltech-256~\cite{caltech_256}. We randomly select 20,000 images from each dataset and split them into two subsets: 10,000 samples detection dataset and 10,000 samples clean dataset. All images are resized to a consistent 256x256 resolution. However, for diffusion model-specific watermarks, we utilize the image prompt datasets DiffusionDB~\cite{wang2022diffusiondb}, MidJourney Prompt dataset~\cite{midjounrydataset} and image captions from COCO dataset~\cite{lin2014coco}. Similarly, we randomly choose 20,000 text prompts and divide them into two splits. We employ Stable Diffusion V1.4~\cite{rombach2022high} as the image generation model, maintaining the generated image size at 256x256. 

Regarding the detection model, $\AlgName$ is designed as a watermark detection method, allowing any DNN model to be plugged in as the detection model. For efficiency, we construct a simple network consisting of only 5 ConvNext-V2 blocks~\cite{convnext_v2} with only 1.93M parameters.

\subsection{Watermark Detection}

In this section, we will test $\AlgName$ detection performance across multiple dataset and watermark method, our evaluation will have two parts, single watermark and multiple watermarks.

\begin{table}[!b]
\vspace{-1.0em}
\caption{Watermark detection performance of $\AlgName$ across different datasets and watermark methods. Methods marked with $^*$ represent generative watermarks, which are directly embedded during the image generation process. Higher AUC and TPR @ 0.1 FPR indicate better performance, while lower FPR @ 0.9 TPR is desirable.}
\label{tab:post_signle_experiments}
\resizebox{1.0\textwidth}{!}{
\begin{tabular}{c|ccc|ccc|ccc}
\hline
       & \multicolumn{3}{c|}{ImageNet}                                                                                                                                         & \multicolumn{3}{c|}{COCO}                                                                                                                                             & \multicolumn{3}{c}{Caltech}                                                                                                                                           \\ \hline
       & \multicolumn{1}{c|}{AUC ($\uparrow$)}   & \multicolumn{1}{c|}{\begin{tabular}[c]{@{}c@{}}TPR ($\uparrow$) @ \\ 0.1 FPR\end{tabular}} & \begin{tabular}[c]{@{}c@{}}FPR ($\downarrow$) @ \\ 0.9 TPR\end{tabular} & \multicolumn{1}{c|}{AUC ($\uparrow$)}   & \multicolumn{1}{c|}{\begin{tabular}[c]{@{}c@{}}TPR ($\uparrow$) @ \\ 0.1 FPR\end{tabular}} & \begin{tabular}[c]{@{}c@{}}FPR ($\downarrow$) @ \\ 0.9 TPR\end{tabular} & \multicolumn{1}{c|}{AUC ($\uparrow$)}   & \multicolumn{1}{c|}{\begin{tabular}[c]{@{}c@{}}TPR ($\uparrow$) @ \\ 0.1 FPR\end{tabular}} & \begin{tabular}[c]{@{}c@{}}FPR ($\downarrow$) @ \\ 0.9 TPR\end{tabular} \\ \hline
LSB    & \multicolumn{1}{c|}{0.852} & \multicolumn{1}{c|}{0.742}                                                    & 0.145                                                    & \multicolumn{1}{c|}{0.837} & \multicolumn{1}{c|}{0.710}                                                    & 0.156                                                    & \multicolumn{1}{c|}{0.845} & \multicolumn{1}{c|}{0.693}                                                    & 0.102                                                    \\ \hline
DDS    & \multicolumn{1}{c|}{0.968} & \multicolumn{1}{c|}{0.938}                                                    & 0.020                                                    & \multicolumn{1}{c|}{0.961} & \multicolumn{1}{c|}{0.927}                                                    & 0.070                                                    & \multicolumn{1}{c|}{0.955} & \multicolumn{1}{c|}{0.941}                                                    & 0.080                                                    \\ \hline
HiDDeN & \multicolumn{1}{c|}{0.952} & \multicolumn{1}{c|}{0.892}                                                    & 0.115                                                    & \multicolumn{1}{c|}{0.954} & \multicolumn{1}{c|}{0.918}                                                    & 0.042                                                    & \multicolumn{1}{c|}{0.944} & \multicolumn{1}{c|}{0.870}                                                    & 0.145                                                    \\ \hline
SS     & \multicolumn{1}{c|}{0.912} & \multicolumn{1}{c|}{0.819}                                                    & 0.082                                                    & \multicolumn{1}{c|}{0.936} & \multicolumn{1}{c|}{0.921}                                                    & 0.102                                                    & \multicolumn{1}{c|}{0.889} & \multicolumn{1}{c|}{0.893}                                                    & 0.094                                                    \\ \hline \hline
     & \multicolumn{3}{c|}{COCO}                                                                                                                                             & \multicolumn{3}{c|}{DiffusionDB}                                                                                                                                      & \multicolumn{3}{c}{MidJourney}                                                                                                                                        \\ \hline
SSig* & \multicolumn{1}{c|}{0.939} & \multicolumn{1}{c|}{0.879}                                                    & 0.118                                                    & \multicolumn{1}{c|}{0.941} & \multicolumn{1}{c|}{0.903}                                                    & 0.045                                                    & \multicolumn{1}{c|}{0.932} & \multicolumn{1}{c|}{0.858}                                                    & 0.147                                                    \\ \hline
TR*   & \multicolumn{1}{c|}{0.821} & \multicolumn{1}{c|}{0.749}                                                    & 0.142                                                    & \multicolumn{1}{c|}{0.823} & \multicolumn{1}{c|}{0.782}                                                    & 0.189                                                    & \multicolumn{1}{c|}{0.811} & \multicolumn{1}{c|}{0.704}                                                    & 0.108                                                    \\ \hline
\end{tabular}}
\vspace{-1.0em}
\end{table}

\noindent \textbf{Single Watermark.} In this set of experiments, we consider the scenario where only one type of watermark method is applied to the detection dataset. As discussed in Section \ref{sec:watermark_ratio}, we randomly watermarked 5\% of the images in the detection dataset using a single watermark method.

The upper part of Table \ref{tab:post_signle_experiments} presents the results for post-processing watermarks. $\AlgName$ achieves remarkable detection performance across all methods and datasets, with AUC scores consistently above 0.8. However, the performance on the LSB watermark is relatively lower compared to other methods. This can be attributed to the fact that LSB watermarks introduce minimal modifications to the image, as evidenced by their lower PSNR values (see Appendix~\ref{sec:visualization}).
The lower part of Table \ref{tab:post_signle_experiments} shows the results for generative watermarks. $\AlgName$ maintains AUC scores above 0.9 for the Stable Signature (SSig) watermark across all datasets. However, the detection rates for the Tree-Ring (TR) watermark are comparatively lower. This may be due to the fact that the changes brought by TR are smaller than other watermarking methods since it embeds the watermark into the diffusion latent space.

\begin{table}[!th]
\vspace{-1.25em}
\caption{Watermark detection performance of $\AlgName$ across different datasets and watermark methods in a multi-watermark setting. For post-processing watermarks, each method is applied to 1.25$\%$ of the dataset, resulting in a total of 5$\%$ watermarked images. Methods marked with $^*$ represent generative watermarks, where each method is applied to 2.5$\%$ of the dataset during the image generation process, also resulting in a total of 5$\%$ watermarked images.
}
\label{tab:post_multi_experments}
\vspace{-10pt}
\resizebox{1.0\textwidth}{!}{
\begin{tabular}{c|ccc|ccc|ccc}
\hline
       & \multicolumn{3}{c|}{ImageNet}                                                                                                                                         & \multicolumn{3}{c|}{COCO}                                                                                                                                             & \multicolumn{3}{c}{Caltech}                                                                                                                                           \\ \hline
       & \multicolumn{1}{c|}{AUC ($\uparrow$)}   & \multicolumn{1}{c|}{\begin{tabular}[c]{@{}c@{}}TPR ($\uparrow$) @ \\ 0.1 FPR\end{tabular}} & \begin{tabular}[c]{@{}c@{}}FPR ($\downarrow$) @ \\ 0.9 TPR\end{tabular} & \multicolumn{1}{c|}{AUC ($\uparrow$)}   & \multicolumn{1}{c|}{\begin{tabular}[c]{@{}c@{}}TPR ($\uparrow$) @ \\ 0.1 FPR\end{tabular}} & \begin{tabular}[c]{@{}c@{}}FPR ($\downarrow$) @ \\ 0.9 TPR\end{tabular} & \multicolumn{1}{c|}{AUC ($\uparrow$)}   & \multicolumn{1}{c|}{\begin{tabular}[c]{@{}c@{}}TPR ($\uparrow$) @ \\ 0.1 FPR\end{tabular}} & \begin{tabular}[c]{@{}c@{}}FPR ($\downarrow$) @ \\ 0.9 TPR\end{tabular} \\ \hline
LSB    & \multicolumn{1}{c|}{0.732} & \multicolumn{1}{c|}{0.642}                                                    & 0.245                                                    & \multicolumn{1}{c|}{0.715} & \multicolumn{1}{c|}{0.595}                                                    & 0.256                                                    & \multicolumn{1}{c|}{0.725} & \multicolumn{1}{c|}{0.573}                                                    & 0.202                                                    \\ \hline
DDS    & \multicolumn{1}{c|}{0.898} & \multicolumn{1}{c|}{0.828}                                                    & 0.120                                                    & \multicolumn{1}{c|}{0.911} & \multicolumn{1}{c|}{0.847}                                                    & 0.170                                                    & \multicolumn{1}{c|}{0.905} & \multicolumn{1}{c|}{0.831}                                                    & 0.180                                                    \\ \hline
HiDDeN & \multicolumn{1}{c|}{0.842} & \multicolumn{1}{c|}{0.782}                                                    & 0.155                                                    & \multicolumn{1}{c|}{0.844} & \multicolumn{1}{c|}{0.778}                                                    & 0.142                                                    & \multicolumn{1}{c|}{0.834} & \multicolumn{1}{c|}{0.737}                                                    & 0.145                                                    \\ \hline
SS     & \multicolumn{1}{c|}{0.802} & \multicolumn{1}{c|}{0.709}                                                    & 0.182                                                    & \multicolumn{1}{c|}{0.826} & \multicolumn{1}{c|}{0.811}                                                    & 0.202                                                    & \multicolumn{1}{c|}{0.779} & \multicolumn{1}{c|}{0.783}                                                    & 0.194                                                    \\ \hline \hline
     & \multicolumn{3}{c|}{COCO}                                                                                                                                             & \multicolumn{3}{c|}{DiffusionDB}                                                                                                                                      & \multicolumn{3}{c}{MidJourney}                                                                                                                                        \\ \hline
SSig* & \multicolumn{1}{c|}{0.829} & \multicolumn{1}{c|}{0.769}                                                    & 0.218                                                    & \multicolumn{1}{c|}{0.831} & \multicolumn{1}{c|}{0.793}                                                    & 0.145                                                    & \multicolumn{1}{c|}{0.822} & \multicolumn{1}{c|}{0.748}                                                    & 0.247                                                    \\ \hline
TR*   & \multicolumn{1}{c|}{0.711} & \multicolumn{1}{c|}{0.639}                                                    & 0.242                                                    & \multicolumn{1}{c|}{0.713} & \multicolumn{1}{c|}{0.672}                                                    & 0.289                                                    & \multicolumn{1}{c|}{0.701} & \multicolumn{1}{c|}{0.594}                                                    & 0.208                                                    \\ \hline
\end{tabular}}
\vspace{-1.25em}
\end{table}

\noindent \textbf{Multiple Watermarks.} In this set of experiments, we evaluate the performance of $\AlgName$ when multiple watermark methods are simultaneously present in the dataset. This scenario more closely resembles real-world conditions where different watermarking techniques may appear in the same dataset. For post-processing watermarks (LSB, DDS, HiDDeN, and SS), we randomly apply each method to 1.25\% of the images in the detection dataset, resulting in a total of 5\% watermarked images. For generative model watermarks (SSig and TR), each method is applied to 2.5\% of the images, again resulting in a total of 5\% watermarked images.

The upper part of Table~\ref{tab:post_multi_experments} presents the results for post-processing watermarks in the multi-watermark setting. Compared to the single watermark scenario, the detection performance of $\AlgName$ decreases slightly across all metrics and datasets. This is expected, as the presence of multiple watermark types introduces additional variability and complexity.
The lower part of Table \ref{tab:post_multi_experments} shows the results for generative watermarks, where each watermark method (SSig and TR) is applied to 2.5\% of the images. The AUC scores for both SSig and TR watermarks are lower compared to the single watermark setting but remain above 0.7. However, the greater decrease in performance for the TR watermark reflects the fact that stealthier watermarks will be further weakened in the presence of multiple watermarks.

\subsection{Ablation Study}
\label{sec:ablation}

\begin{table}[!b]
\centering
\vspace{-10pt}
\caption{Ablation study on the impact of loss function choices for the "Minimize" (clean dataset) and "Maximize" (detection dataset) objectives on watermark detection performance (AUC).}
\label{tab:loss_design}
\vspace{-1.0em}
\resizebox{0.45\textwidth}{!}{
\begin{tblr}{
  cells = {c},
  cell{1}{3} = {c=4}{},
  cell{3}{1} = {r=4}{},
  vline{3} = {1}{},
  vline{3-6} = {2,4-6}{},
  vline{2-6} = {3}{},
  vline{2-6} = {4}{},
  vline{2-6} = {5}{},
  vline{2-6} = {6}{},
  hline{1,3,7} = {-}{},
  hline{2} = {3-6}{},
  hline{4-6} = {2-6}{},
}
                                       &         & Minimize &        &       &                \\
                                       &         & BCE      & Linear & Exp   & Softmax        \\
\begin{sideways}Maximize\end{sideways} & BCE     & 0.757    & 0.612  & 0.831 & 0.877          \\
                                       & Linear  & 0.891    & 0.786  & 0.842 & \textbf{0.968} \\
                                       & Exp     & 0.512    & 0.516  & 0.716 & 0.513          \\
                                       & Softmin & 0.583    & 0.508  & 0.533 & 0.752          
\end{tblr}}
\vspace{-1.25em}
\end{table}

\noindent \textbf{Loss Design.}
The ablation study in \autoref{tab:loss_design} demonstrates the importance of selecting appropriate loss functions for the "Maximize" and "Minimize" objectives in $\AlgName$. Using symmetric loss leads to a performance drop due to the presence of clean samples in the maximizing divert the offset goal. As our analysis in~\autoref{sec:methology}, asymmetric loss design can greatly improve model detection capabilities. The combination of linear loss and softmax loss achieves the best performance (AUC 0.968) by providing a balanced and complementary optimization approach.

\begin{figure*}[!t]
    \centering
    % \vspace{-1.5em}
    \includegraphics[width=\textwidth]{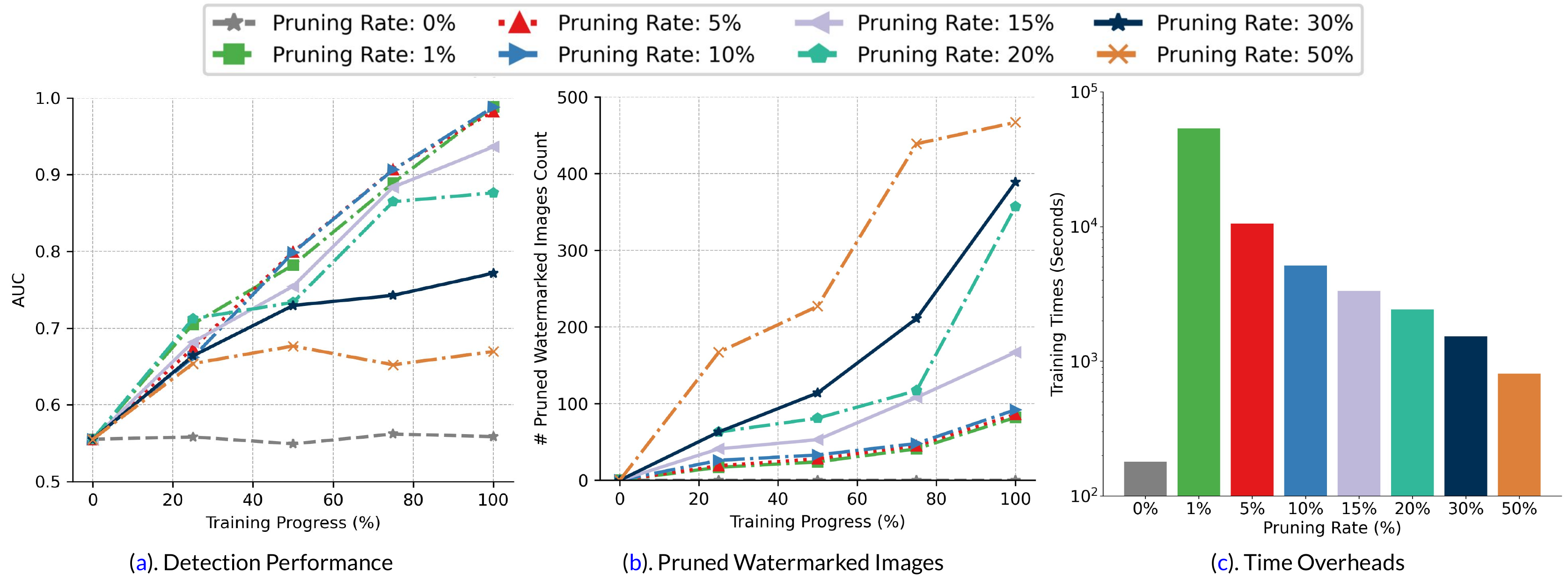}
    \vspace{-1.5em}
    \caption{Impact of pruning rate on watermark detection and training overheads. \textcolor{blue}{\textbf{(a)}} Detection performance measured by AUC decreases as the pruning rate increases, with higher pruning removing more watermarked images during training. \textcolor{blue}{\textbf{(b)}}Number of pruned watermarked images increases with higher pruning rates throughout the training process. \textcolor{blue}{\textbf{(c)}} Time overheads for training increase substantially with higher pruning rates.}
    \vspace{-2.0em}
    \label{fig:ablation_prunerates}
\end{figure*}

\noindent \textbf{Pruning Rate.}
We investigate the impact of the pruning rate on the performance and efficiency of $\AlgName$ and present the results in \autoref{fig:ablation_prunerates}. As shown in \textcolor{blue}{\textbf{(a)}}, excessively high pruning rates result in reduced detection performance because the model has not successfully separated watermarked images from clean images at this stage. Pruning too many images in this scenario will cause a large number of watermarked images to be removed, as shown in \textcolor{blue}{\textbf{(b)}}, thus affecting the model's ability to learn from the remaining watermarked images in the next round of learning. However, while lower pruning rates maintain high detection performance, they require more iterations and incur significant time overhead. As illustrated in \autoref{fig:ablation_prunerates}(c), a 1\% pruning rate incurs approximately 66.2 times more overhead compared to a 50\% pruning rate.

\section{Discussion}

\noindent \textbf{Limitations and Future Work.}
While $\AlgName$ demonstrates strong detectability, our evaluations do reveal some limitations. One key challenge is that the detection performance relies on the clean dataset and the detection dataset having similar distributions, which may be difficult to achieve in practice. Additionally, the current method relies on hyperparameter tuning to work effectively. Future research should focus on addressing these limitations by exploring adaptive techniques such as domain adaptation to handle distribution mismatches or enhance detectability through optimizable hyperparameter selection.

\noindent \textbf{Wider Applications.}
In addition to detecting invisible watermarks, $\AlgName$ demonstrates versatility in supporting a range of other applications. As detailed in Appendix~\autoref{sec:othercase}, $\AlgName$ can be utilized to facilitate watermark removal attacks (Appendix~\autoref{sec:wm_remove}) and to filter out harmful examples from datasets (Appendix~\autoref{sec:harmful_exampls}). These use cases showcase the potential of $\AlgName$ to make a significant impact across various domains, extending beyond its core functionality.

\noindent \textbf{Broaden Impacts.}
By enabling reliable invisible watermark detection, $\AlgName$ allows users to make informed decisions about image usage, deters unauthorized watermarking, and promotes responsible practices. Its implications extend to industries like digital forensics, assisting in identifying image tampering and unauthorized distribution. As invisible watermark usage evolves, $\AlgName$'s impact is poised to grow, fostering transparency, accountability, and trust in digital visual content handling, ultimately contributing to a more secure digital image ecosystem.

\section{Conclusion}
In this paper, we proposed $\AlgName$, a Black-box invisible watermark detection method that achieves robust performance across diverse watermarking techniques without prior knowledge of the specific method used to apply the watermarks. Extensive evaluations demonstrated $\AlgName$'s effectiveness, with AUC scores consistently above 0.9 in most single-watermark settings and above 0.7 in challenging multi-watermark scenarios. $\AlgName$'s potential extends beyond detection to supporting watermark removal attacks and filtering harmful examples. As invisible watermarks become increasingly prevalent, especially in AI-generated imagery, $\AlgName$ capability to identify watermarked samples lays the foundation for promoting transparency, accountability, and trust in digital visual content.

\clearpage  % TODO REVIEW/FINAL: This \clearpage needs to be removed from both review and camera-ready versions.

\bibliographystyle{splncs04}
\bibliography{main}

\clearpage
\appendix

{\centering
\textbf{Finding needles in a haystack: A Black-Box Approach to Invisible Watermark Detection}\\
\vspace{0.5em}Supplementary Material \\}

\section{Detailed Experiments Setting \& Hyperparameters}
\label{sec:Hyperparameters}
In this Section, we will detail the report the experiments setting \& hyperparameters that we used in the \autoref{sec:evaluation}:

\noindent \textbf{Hardware \& Software.} All experiments were conducted on a server equipped with 4 NVIDIA Tesla V100 GPUs, an AMD EPYC 7763 CPU, and 256 GB of RAM. The software environment includes CUDA 12.0 and PyTorch 2.2.1.

\noindent \textbf{Watermarks.} In \autoref{sec:evaluation}, we use a series of the watermark method to test the performance of $\AlgName$, we report their setting and hyperparameters in Table~\ref{tab:watermark_hyperparameters}. 

\begin{table}
\centering
\caption{Hyperparameter settings for different watermarking methods used in the experiments.}
\begin{tblr}{
cell{3}{1} = {r=2}{},
cell{3}{2} = {r=2}{},
cell{5}{1} = {r=4}{},
cell{5}{2} = {r=4}{},
cell{9}{1} = {r=6}{},
cell{9}{2} = {r=6}{},
cell{16}{1} = {r=2}{},
cell{16}{2} = {r=2}{},
hline{2-3,5,9,15-16, 18},
hline{2} = {1.3pt},
}
\textbf{Watermark} & \textbf{Embedded Bits} & \textbf{Parameters} & \textbf{Values} \\
LSB~\cite{lsb2010lsbwatermark} & 64 bits & NA & NA \\
DctDwtSVD~\cite{4554423} & 64 bits & Scales (Y, U, V) & 0, 36, 0 \\
& & Block & 64 \\
HiDDeN~\cite{zhu2018hidden} & 64 bits & Crop & 0.2-0.25 \\
& & Cropout & 0.55-0.6 \\
& & Dropout & 0.55-0.6 \\
& & JPEG & 0.8 \\
StegaStamp~\cite{tancik2020stegastamp} & 64 bits & Brightness & 0.3 \\
& & Random Noise & 0.02 \\
& & Saturation & 1.0 \\
& & Hue & 0.1 \\
& & Contrast & 0.5-1.5 \\
& & JPEG & 0.5 \\
StableSignature~\cite{fernandez2023stable} & 64 bits & Diffusion Model & Stable Diffusion V1.4 \\
Tree-Ring~\cite{wen2023tree} & NA & Diffusion Model & Stable Diffusion V1.4 \\
& & Type & $\text{Tree-Ring}_{\textbf{Rings}}$~
\end{tblr}
\label{tab:watermark_hyperparameters}
\end{table}

\noindent \textbf{Detection.} We will report the hyperparameters and settings used for $\AlgName$ in \autoref{sec:evaluation}. The results are presented in Table~\ref{tab:detection_hyperparameters}.

\begin{table}
\centering
\caption{Hyperparameter settings for the $\AlgName$ watermark detection method.}
\begin{tabular}{l|l}
\textbf{Parameters} & \textbf{Values} \\
\hline
Optimizer & AdamW \\
Base Learning Rate & 1e-4 \\
Weight Decay & 0.01 \\
Momentum & $\beta$1,$\beta$2 = 0.9,0.999 \\
Batch Size & 32 \\
Training Epochs & 50 \\
Pruning Rate & 0.10 \\
Pruning Interval & 10
\end{tabular}
\label{tab:detection_hyperparameters}
\end{table}

\section{Wider Applications}
\label{sec:othercase}
\subsection{Watermark removal attacks}
\label{sec:wm_remove}
Several watermark removal attacks~\cite{zhao2023invisible,lukas2023leveraging, saberi2023robustness, jiang2023evading} have been proposed to remove or break the watermark in an image, causing the decoding algorithm to fail. These methods can be classified into two categories: black-box and white-box. Black-box methods, such as Regenerate~\cite{zhao2023invisible}, employ a diffusion model to regenerate the watermarked image, hoping that the prior knowledge from the diffusion model will remove the watermark. However, such methods have limited removal performance and fail when attempting to remove advanced watermarks like StegaStamp~\cite{tancik2020stegastamp} or Tree-Ring~\cite{wen2023tree}. White-box algorithms, such as Warfare~\cite{li2023towards}, WEvade-B-S~\cite{jiang2023evading}, and SurPGD~\cite{saberi2023robustness}, require a pure watermarked dataset and a clean dataset to train a surrogate model and perform PGD~\cite{madry2017towards} adversarial attacks on the surrogate model to achieve watermark removal. However, obtaining a pure watermarked dataset in real life is challenging, as discussed in \autoref{sec:wm_detection}.

$\AlgName$ bridges the gap between white-box and black-box removal attacks. As $\AlgName$ can detect watermarks in real-world, watermark-blended detection datasets, we can use the detection results to form a relatively pure watermarked dataset, enabling downstream black-box watermark removal attacks. We present the results in \autoref{tab:pgd_attack}. In this evaluation, we increase the dataset size to 50,000 while maintaining the watermark ratio at 5\%. The PGD attack settings are kept the same as in SurPGD~\cite{saberi2023robustness}.

\begin{table}
\centering
\caption{PGD watermark removal attack results using $\AlgName$ for watermark detection. Lower AUC scores indicate better removal performance.}
\begin{tblr}{
  cells = {c},
  vline{2} = {-}{},
  hline{2} = {-}{},
}
    & HiDDeN & SS    & SSig  & TR    \\
AUC$\downarrow$ & 0.572  & 0.613 & 0.552 & 0.504 
\end{tblr}
\label{tab:pgd_attack}
\end{table}

The results demonstrate that by leveraging $\AlgName$'s watermark detection capabilities to create a relatively pure watermarked dataset, the effectiveness of black-box watermark removal attacks can be significantly improved. The lower AUC scores indicate that the PGD attack, guided by $\AlgName$'s detection, achieves better removal performance across various watermarking methods, including HiDDeN~\cite{zhu2018hidden}, StegaStamp (SS)~\cite{tancik2020stegastamp}, StableSignature (SSig)~\cite{fernandez2023stable}, and Tree-Ring (TR)~\cite{wen2023tree}. This highlights the potential of $\AlgName$ to support and enhance watermark removal attacks in real-world scenarios where pure watermarked datasets are not readily available.

\subsection{Filtering invisible anomalies}
\label{sec:harmful_exampls}
Besides watermarks, numerous techniques have been proposed to insert invisible information into image datasets for various purposes, such as data poisoning~\cite{li_ISSBA_2021, zeng2021rethinking}, tracing dataset usage~\cite{wang2023detect}, or preventing models from learning specific information~\cite{shan2023glaze, sandoval2022autoregressive}. As our framework formulation in \autoref{sec:methology} shows, $\AlgName$ is capable of detecting any data that does not follow the same distribution as clean data. This naturally raises the question: can $\AlgName$ also detect these special samples in the dataset? To investigate this, we select representative works from each category. For data poisoning, we choose SSBA~\cite{li_ISSBA_2021}, which uses an autoencoder to generate backdoor samples and then poisons the downstream model to predict the samples to a target class with a pre-defined backdoor trigger. Another data poisoning method is Frequency~\cite{zeng2021rethinking}, which embeds frequency-adaptive noise into the dataset to achieve a backdoor attack while evading frequency-based detection. For dataset tracing, we use DIAGNOSIS~\cite{wang2024diagnosis}, which inserts imperceptible distortions into images. The downstream trained diffusion model learns these distortions and generates examples with similar distortions, enabling the tracing of the model's training dataset source. Other methods include unlearnable examples, such as AR~\cite{sandoval2022autoregressive}, which inserts autoregressive noise into images, causing the downstream classifier to focus on learning the noise rather than the image features. This protects the visual information in the image and leads to poor model performance on clean images. GLAZE~\cite{shan2023glaze} inserts invisible optimized noise into images, causing diffusion models to fail to learn the visual information and protecting artists' work from being stolen by diffusion models. We maintain the same settings as the evaluation in \autoref{sec:evaluation} and adjust the insertion ratio to match their original work. The results are presented in \autoref{tab:other_samples}.

\begin{table}
\centering
\caption{Detection results for various types of invisible information inserted into image datasets, along with their respective insertion ratios.}
\resizebox{0.85\columnwidth}{!}{
\begin{tblr}{
  colspec = {X[c]X[c]X[c]X[c]X[c]},
  cells = {c},
  cell{1}{2} = {c=2}{},
  cell{1}{5} = {c=2}{},
  vline{2-5} = {1}{},
  vline{2-6} = {2-5}{},
  hline{1,3-6} = {-}{},
  hline{2} = {2-6}{},
}
                 & Data Poisoning &                   & Dataset Tracing   & Unlernable Example &               \\
                 & {SSBA\\10\%}   & {Frequency\\10\%} & {DIAGNOSIS\\20\%} & {AR\\10\%}         & {GLAZE\\25\%} \\
AUC ($\uparrow$)              & 0.987          & 0.975             & 0.653             & 0.903              & 0.934         \\
{TPR ($\uparrow$) @\\0.1 FPR} & 0.944          & 0.931             & 0.554             & 0.823              & 0.856         \\
{FPR ($\downarrow$) @\\0.9 TPR} & 0.010          & 0.016             & 0.273             & 0.048              & 0.014         
\end{tblr}}
\label{tab:other_samples}
\end{table}

The evaluation results demonstrate that $\AlgName$ is highly effective in detecting various types of invisible information inserted into image datasets. It achieves high AUC scores for data poisoning methods like SSBA and Frequency, as well as for unlearnable examples such as AR and GLAZE. However, $\AlgName$'s performance on dataset tracing using DIAGNOSIS is relatively lower, suggesting that the imperceptible distortions inserted by this method may be more challenging to detect. Overall, the results highlight the versatility and effectiveness of $\AlgName$ in identifying a wide range of invisible information in image datasets, demonstrating its potential as a powerful tool for ensuring dataset integrity and protecting against malicious manipulations.

\subsection{Synthesis clean data}
\label{sec:systhesis_clean_data}

In some scenarios, obtaining a clean dataset to serve as a reference for $\AlgName$ can be challenging, which may hinder the effectiveness of the watermark detection process. To mitigate the challenge of obtaining clean data in some scenarios, we explore the use of synthesized data to extend the clean dataset. We choose the DctDwtSvd~\cite{4554423} watermarking method, set the watermark ratio to 5\%, and use a detection dataset size of 10,000. We then evaluate several synthesis methods:

\begin{packeditemize}
\item \textbf{Case 1, Detection Dataset to Text to Image:} In this case, we use BLIP-2~\cite{li2023blip} as the caption model to generate captions for the detection dataset. Since the watermarks are invisible, we expect the generated captions to not contain watermark information. We then use these generated prompts to feed the generation model and synthesize high-quality clean images for the clean dataset.

\item \textbf{Case 2, Clean Dataset to Text to Image:} This case follows the same setting as Case 1, but the captions are generated from the clean dataset. As the amount of real clean data may be less than 50\% of the total clean dataset, we use the same prompt multiple times with different random seeds to generate different images.

\item \textbf{Case 3, Clean Dataset Image Variation:} This case has the same settings as \textbf{Case 2}, but instead of using prompts to generate the synthesized images, we directly use the clean images as the condition to guide the generation~\cite{varition} and obtain variations of the clean data.

\item \textbf{Case 4, Synthesized Text Prompt to Image:} In this case, we use the captions from \textbf{Case 2} and input them into ChatGPT~\cite{achiam2023gpt} to generate variations of these prompts. We then use these prompts to synthesize clean images.
\end{packeditemize}

For all generation models, we use Stable Diffusion V2~\cite{rombach2022high} with a step size of 100, keeping all other hyperparameters at their default values. The results are shown in Table~\ref{tab:synthesis_clean_data}.

\begin{table}
\centering
\caption{AUC scores for different ratios of synthesized data in the clean dataset. The results demonstrate the impact of using synthesized data on watermark detection performance.}
\begin{tblr}{
  cells = {c},
  cell{2}{2} = {r=4}{},
  vline{2-6} = {1-2}{},
  vline{2,3-6} = {3-5}{},
  hline{1-2,6} = {-}{},
  hline{3-5} = {1,3-6}{},
}
       & 0\%   & 20\%   & 40\%   & 60\%   & 80\%   \\
Case 1 & 0.968 & 0.966 & 0.958 & 0.784 & 0.684 \\
Case 2 &       & 0.942 & 0.911 & 0.745 & 0.689 \\
Case 3 &       & 0.910 & 0.886 & 0.613 & 0.500 \\
Case 4 &       & 0.940 & 0.924 & 0.734 & 0.702 
\end{tblr}
\label{tab:synthesis_clean_data}
\end{table}

The evaluation results show that using synthesized data to extend the clean dataset can be effective, but the performance degrades as the ratio of synthesized data increases. Case 1, which generates captions from the detection dataset and uses them to synthesize clean images, achieves the best performance among the synthesis methods. This suggests that the captions generated from the detection dataset capture relevant content while excluding watermark information. Cases 2 and 4, which use captions from the clean dataset or variations of those captions, also demonstrate good performance, although slightly lower than Case 1. Case 3, which directly uses image variations of the clean data, shows the lowest performance among the synthesis methods, indicating that image-based variations may introduce more noise or artifacts that affect watermark detection. Overall, these results highlight the potential of using synthesized data to mitigate the lack of clean data, but the ratio of synthesized data should be carefully considered to maintain high watermark detection performance.

\section{Further Ablation Study}

\begin{table}
\centering
\caption{Watermark detection performance (AUC) across different combinations of clean (reference) datasets and detection (watermarked) datasets. Diagonal elements represent same-domain scenarios, while off-diagonal elements represent cross-domain scenarios. Watermark detection is most effective when the clean and detection datasets are from the same domain, with some domains generalizing better than others in cross-domain cases.}
\resizebox{0.90\columnwidth}{!}{
\begin{tblr}{
  row{even} = {c},
  row{1} = {c},
  row{5} = {c},
  cell{1}{3} = {c=6}{},
  cell{3}{1} = {r=3}{},
  cell{3}{2} = {c},
  cell{3}{3} = {c},
  cell{3}{4} = {c},
  cell{3}{5} = {c},
  cell{3}{6} = {c},
  cell{3}{7} = {c},
  cell{3}{8} = {c},
  vline{3} = {1}{},
  vline{3-8} = {2,4-5}{},
  vline{2-8} = {3}{},
  vline{2-8} = {4}{},
  vline{2-8} = {5}{},
  hline{1-3,6} = {-}{},
  hline{4-5} = {2-8}{},
}
                                        &             & Clean    &       &             &       &             &               \\
\begin{sideways}\end{sideways}          &             & ImageNet & COCO  & Caltech-256 & CelebA & iNaturalist & Stanford Cars \\
\begin{sideways}Detection\end{sideways} & ImageNet    & 0.968    & 0.957 & 0.942       & 0.612 & 0.909       & 0.632         \\
                                        & COCO        & 0.948    & 0.961 & 0.916       & 0.607 & 0.914       & 0.658         \\
                                        & Caltech-256 & 0.916    & 0.937 & 0.955       & 0.599 & 0.893       & 0.686         
\end{tblr}}
\label{tab:mismatch_dataset}
\end{table}

\noindent \textbf{Mismatch Dataset.}
The ablation study in \autoref{tab:mismatch_dataset} investigates the impact of domain mismatch between the clean (reference) and detection (watermarked) datasets on the performance of $\AlgName$. The highest detection performance, with AUC scores above 0.9, is achieved when both datasets are from the same domain (diagonal elements). Cross-domain scenarios (off-diagonal elements) exhibit varying performance, with some combinations, like ImageNet and COCO, showing strong generalization (AUC 0.957), while others, like Celeb and other domains, have limited effectiveness (AUC 0.599 to 0.612).

The study highlights the importance of similarity between the clean and detection datasets for optimal performance. Domain mismatch can degrade the model's ability to generalize and learn distinguishing features. To mitigate this, it is recommended to use a clean dataset that closely matches the detection dataset or exhibits strong generalization, such as ImageNet. Further research could explore domain adaptation techniques and incorporating diverse clean images to improve robustness to dataset mismatch.

\begin{figure*}[!t]
    \centering
    % \vspace{-1.5em}
    \includegraphics[width=\textwidth]{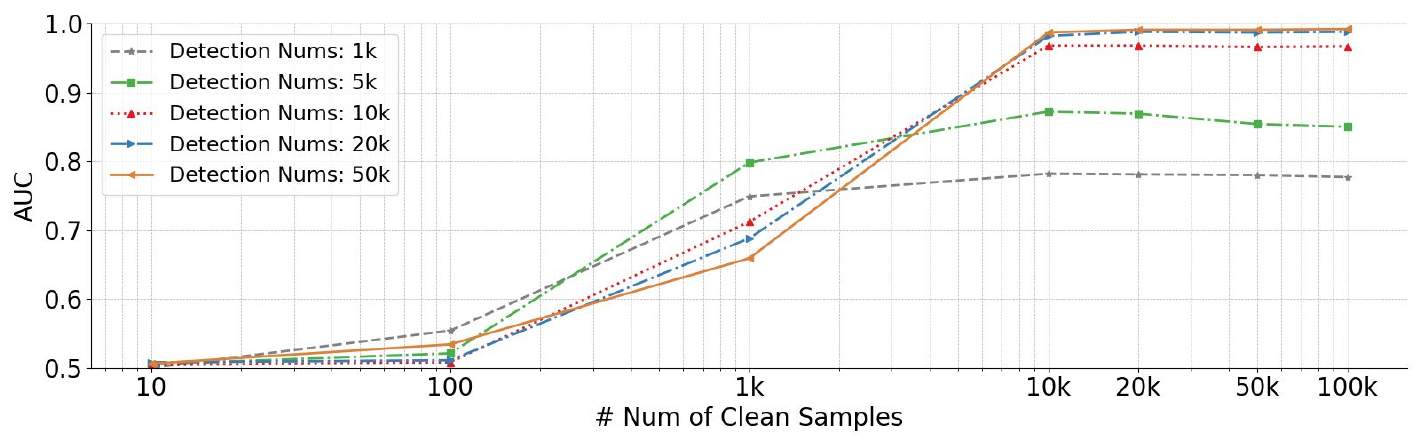}
    \vspace{-.5em}
    \caption{Watermark detection performance (measured by AUC) with varying sizes of the clean reference dataset and different numbers of clean samples used during training. Larger reference datasets and more clean training samples generally lead to better detection performance, with diminishing returns after a certain point.}
    \vspace{-1em}
    \label{fig:clean_samples}
\end{figure*}

\noindent \textbf{Clean-set Size.}
The ablation study in \autoref{fig:ablation_prunerates} investigates the impact of the clean reference dataset size and the number of clean samples used during training on the watermark detection performance of $\AlgName$. The results show that increasing the dataset size and the number of clean training samples leads to better detection performance, with AUC scores consistently exceeding 0.9 for the largest dataset size (50,000 samples). However, the performance gains exhibit diminishing returns beyond a certain point, suggesting a trade-off between computational cost and marginal improvements. The study also reveals that using a small detection dataset can limit the detection performance due to overfitting and the inability to generalize to the full watermark distribution. To mitigate this issue, it is crucial to ensure that both the clean reference dataset and the detection dataset are sufficiently large and diverse to capture the variability in watermark patterns and image characteristics, enabling the model to learn robust and generalizable features for effective watermark detection.

\section{Discussion}

\noindent \textbf{Relation to Steganalysis.} Steganalysis~\cite{boroumand2018deep, kheddar2024deep} refers to methods that detect secret messages embedded in digital media using steganography. Given the similarity between steganography and watermarking, $\AlgName$ may remind people of steganalysis. Although both watermarking and steganography involve embedding information into digital media, they serve different purposes. Watermarking aims to protect intellectual property rights and ensure the authenticity of digital content by embedding a unique identifier or signature. The embedded watermark is typically designed to be robust against various image processing operations and attacks, and its presence should be detectable even if the image undergoes modifications. On the other hand, steganography focuses on concealing the existence of a hidden message within digital media, prioritizing undetectability over robustness. The goal is to communicate secretly without raising suspicion, and the success of steganography relies on the inability of an adversary to distinguish between normal and steganographic media.

Besides the difference between steganography and watermarking, the detection assumptions and goals of $\AlgName$ and steganalysis are different. Steganalysis requires a paired dataset, where the original image (cover) and its steganographic version (stego) are provided for analysis and detection. This means that steganalysis relies on knowledge of the specific steganography algorithm and access to the original version of the image. In contrast, $\AlgName$ does not require such knowledge or the original version of the watermarked image. Furthermore, the detection goals of steganalysis and $\AlgName$ differ. Steganalysis focuses on training on a labeled dataset and aims to generalize to a testing dataset, ensuring that both datasets use the same steganography method. The success of steganalysis relies on the consistency of the steganography algorithm across the training and testing data. On the other hand, $\AlgName$ operates in a more realistic and challenging scenario where there is no separate training and testing dataset. Instead, $\AlgName$ performs detection solely on the given detection dataset, aiming to split it into watermarked and clean parts without prior knowledge of the watermarking algorithm or access to labeled data. This makes $\AlgName$ more flexible and applicable to real-world situations where the watermarking method may be unknown, and labeled data is unavailable. While steganalysis relies on the consistency of the steganography algorithm between the training and testing data, $\AlgName$ can handle the presence of multiple, unknown watermarking techniques within a single dataset. This makes $\AlgName$ a more versatile tool for detecting watermarks in real-world scenarios where the watermarking methods may be diverse and unknown.

In summary, although $\AlgName$ and steganalysis share the goal of detecting hidden information in digital media, they differ in their assumptions, requirements, and detection goals.

\noindent \textbf{Safety \& Security Concern.} While $\AlgName$ demonstrates significant potential in detecting invisible watermarks, it is essential to address the safety and security concerns that may arise from its use. One major concern is the possibility of $\AlgName$ leaking secret information embedded in the watermarks. As $\AlgName$ is designed to detect the presence of watermarks, it may inadvertently expose sensitive data, such as copyright information, ownership details, or hidden messages, to unauthorized parties. This could compromise the privacy and security of the watermark owners and the intended recipients of the embedded information.

Another concern is the potential misuse of $\AlgName$ to support watermark removal attacks. As discussed earlier, $\AlgName$ can be used to create a relatively pure watermarked dataset, which can then be exploited to train models for watermark removal. This may encourage malicious actors to use $\AlgName$ to circumvent copyright protection and remove watermarks from digital content without permission. Such actions could undermine the effectiveness of watermarking as a security measure and lead to the infringement of intellectual property rights.

To mitigate these concerns, it is crucial to develop safeguards and responsible usage guidelines for $\AlgName$. One approach could be to incorporate a mechanism that prevents the extraction or decoding of the actual watermark information, ensuring that only the presence of watermarks is detected without revealing the embedded data. Additionally, implementing access controls and authentication measures could help restrict the use of $\AlgName$ to authorized parties and prevent its misuse for malicious purposes.

\noindent \textbf{Law \& Policy Implications.} The development and use of $\AlgName$ for invisible watermark detection raise important legal and policy considerations. As watermarking plays a crucial role in protecting intellectual property rights and ensuring the authenticity of digital content, the ability to detect and identify watermarks has significant implications for copyright law and digital rights management.

From a legal perspective, $\AlgName$ could serve as a valuable tool for copyright holders to enforce their rights and detect unauthorized use of their watermarked content. By enabling the detection of invisible watermarks, $\AlgName$ can help identify instances of copyright infringement and provide evidence for legal action. This could strengthen the position of copyright holders and deter potential infringers from misusing watermarked content.

However, the use of $\AlgName$ also raises concerns about privacy and the potential for abuse. If $\AlgName$ falls into the wrong hands, it could be used to illegally remove watermarks from copyrighted content, facilitating unauthorized distribution and use. This could undermine the effectiveness of watermarking as a copyright protection measure and lead to financial losses for content creators and owners.

To address these concerns, policymakers may need to consider updating existing copyright laws and regulations to account for the emergence of advanced watermark detection techniques like $\AlgName$. This could involve clarifying the legal status of watermark detection tools, defining the permissible uses of such tools, and establishing penalties for their misuse.

\section{Visualization}
\label{sec:visualization}
\begin{figure*}[!t]
    \centering
    \includegraphics[width=\textwidth]{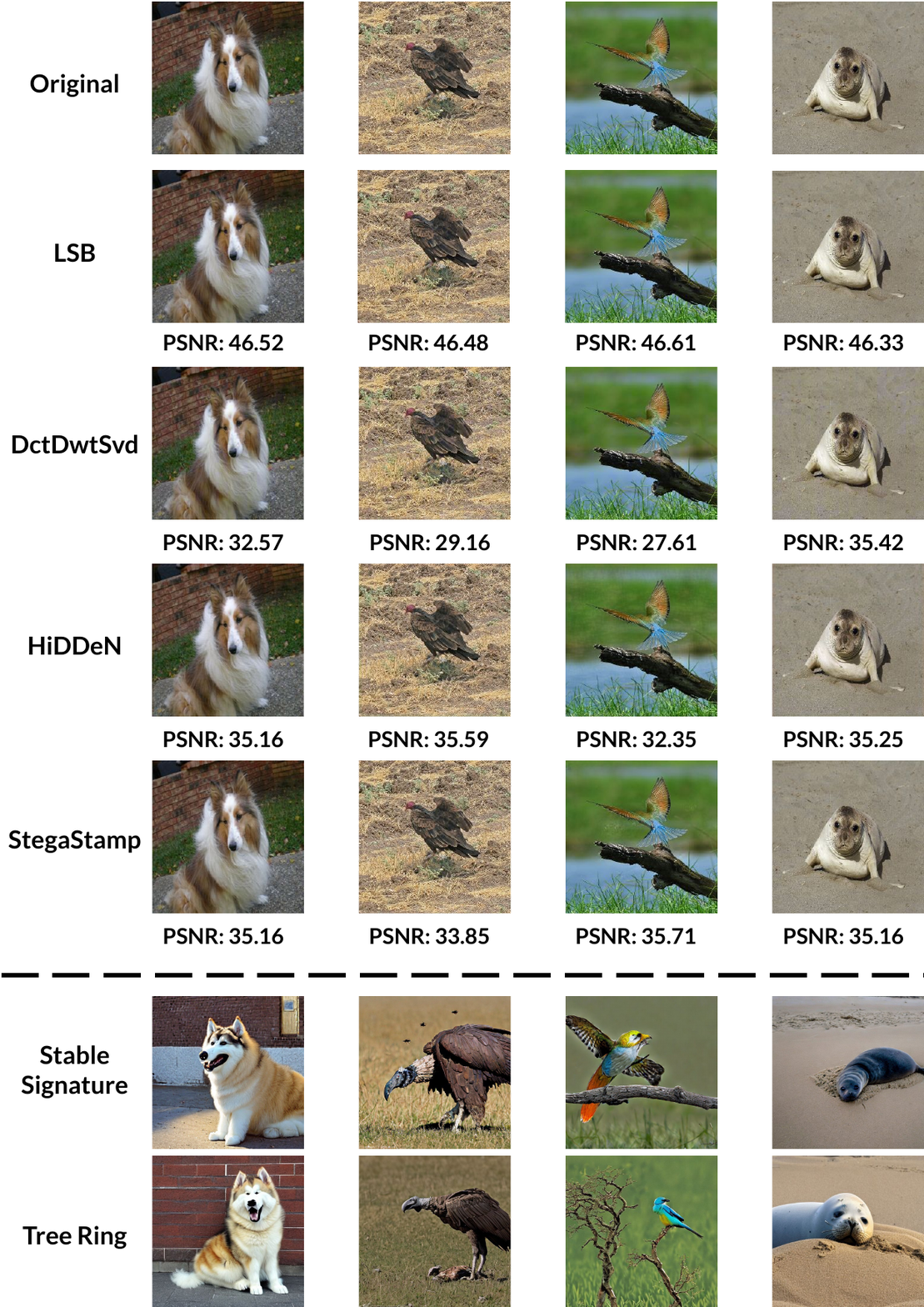}
    \caption{Visual examples of original images and their watermarked counterparts using different watermarking methods. The top row shows the original images. The PSNR values are provided for each post-processing watermarked image, lower PSNR indicating the higher level of distortion introduced by the watermarking process.}
    \label{fig:clean_samples}
\end{figure*}

\end{document}